%% file: root.tex
\pdfoutput=1
\documentclass[a4paper,conference]{IEEEtran}
\usepackage{color}
\usepackage{amsfonts}
\usepackage{subcaption}
\usepackage{multirow}
\usepackage[pdftex]{graphicx}
\ifCLASSINFOpdf
  \usepackage[pdftex]{graphicx}
\else
  \usepackage[dvips]{graphicx}
\fi
\usepackage{caption}
\usepackage{footnote}

\begin{document}
\input{macros}


\title{Augmented Cyclic Consistency Regularization for Unpaired Image-to-Image Translation}

\author{
Takehiko Ohkawa\IEEEauthorrefmark{1},
Naoto Inoue\IEEEauthorrefmark{1}, 
Hirokatsu Kataoka\IEEEauthorrefmark{2},
Nakamasa Inoue\IEEEauthorrefmark{3} \\

\IEEEauthorblockA{
\IEEEauthorrefmark{1}The University of Tokyo, Japan, \\
\IEEEauthorrefmark{2}National Institute of Advanced Industrial Science and Technology (AIST), Tsukuba, Ibaraki, Japan,\\
\IEEEauthorrefmark{3}Tokyo Institute of Technology, Japan.\\
Email: ohkawa-t@iis.u-tokyo.ac.jp,  inoue@hal.t.u-tokyo.ac.jp,
hirokatsu.kataoka@aist.go.jp,
inoue@c.titech.ac.jp
}
}

\maketitle
\footnote[0]{This work was done when the first author was at the Tokyo Institute of Technology.}

\input{abstract}

\IEEEpeerreviewmaketitle

\input{introduction}

\input{related_work}
\input{proposal}
\input{experiments}
\input{conclusion}

\input{reference_short}
\end{document}

%% file: macros.tex
\DeclareRobustCommand\red{\textcolor{red}}
\DeclareRobustCommand\blue{\textcolor{blue}}
\hyphenation{op-tical net-works semi-conduc-tor}
\newcommand{\hyphen}{\mathchar`-}

\def\fig#1#2{
\begin{figure}[!t]
\centering
\includegraphics[width=0.9\hsize]{fig/#1}
\caption{#2}
\label{#1}
\end{figure}
}

\def\figw#1#2{
\begin{figure*}[!t]
\centering
\includegraphics[width=0.9\hsize]{fig/#1}
\caption{#2}
\label{#1}
\end{figure*}
}

\def\tbldata{
\begin{table*}[!t]
\caption{\textbf{Digit translations.} Direction indicates source $\rightarrow$ target translation direction. We measure the impact of the proposed method, ACCR, with respect to classification accuracy (\%) on fake samples in the target domain by fixed classifiers and compare with the CycleGAN~\cite{cyclegan} baseline and CR-CycleGAN. * indicates statistically significant differences at the level of significance $\alpha=0.05$. For ablation we further experiment using CycleGAN with CR-Fake and CR-Rec.}
\centering\begin{tabular}{l|c|c|c|c}
\hline
Model & M $\rightarrow$ MM & MM  $\rightarrow$ M & M  $\rightarrow$ S & S $\rightarrow$ M\\ \hline
CycleGAN & $97.7 \pm 0.3$ & $92.2 \pm 1.2$ & $47.1 \pm 3.1$ & $28.2 \pm 0.9$\\
CR-CycleGAN & $97.7\pm 0.6$ & $94.3 \pm 0.5$* & $43.7 \pm 4.1$ & $29.6 \pm 0.7$\\ \hline
CR-CycleGAN + CR-Fake (Ours) & $97.7 \pm 0.3$ & $93.8 \pm 0.7$* & $46.3 \pm 4.7$ & \mbox{\boldmath $31.9 \pm 3.0$*}\\
CR-CycleGAN + CR-Rec (Ours) &   $97.6 \pm 1.2$ & $94.2 \pm 1.4$* & $48.5 \pm 3.0$ & $30.5 \pm 1.7$* \\
ACCR-CycleGAN (Ours) & \mbox{\boldmath $98.0 \pm 0.5$} & \mbox{\boldmath $94.5 \pm 0.5$*} & \mbox{\boldmath $51.0 \pm 5.2$} & \mbox{\boldmath $31.9 \pm 1.6$*}\\\hline \hline
\end{tabular}
\label{tbl:data}
\end{table*}
}

\def\tbldist{
\begin{table}[!t]
\caption{\textbf{Feature distance between real and augmented samples.} We report the mean squared error (MSE) at the penultimate layer of a discriminator between test and augmented data.}
\centering\begin{tabular}{l|c|c}
\hline
Feature Distance (MSE) & MNIST test  &  MNIST-M test \\\hline
CycleGAN & $41.3 \pm 13.4$ & $46.4 \pm 4.5$ \\
CR-CycleGAN & \mbox{\boldmath $35.1 \pm 6.1$} & $35.4 \pm 8.3$ \\
ACCR-CycleGAN (Ours) & $36.9 \pm 4.9$ & \mbox{\boldmath $29.6 \pm 8.8$} \\
\hline
\hline
\end{tabular}
\label{tbl:dist}
\end{table}
}

\def\tblmap{
\begin{table}[!t]
\caption{\textbf{Maps} $\leftrightarrow$ \textbf{Aerial photograph.} We evaluate the generation quality for fake samples by mean square error (MSE) with ground-truth images.}
\centering\begin{tabular}{l|c|c}
\hline
Model & Maps $\rightarrow$ Photo & Photo $\rightarrow$ Maps\\\hline
CycleGAN & $0.159$ &\mbox{\boldmath $0.023$} \\
CR-CycleGAN & $0.120$ &\mbox{\boldmath $0.023$} \\
ACCR-CycleGAN (Ours) &\mbox{\boldmath $0.117$} &\mbox{\boldmath $0.023$} \\
\hline
\hline
\end{tabular}
\label{tbl:map}
\end{table}
}

\def\tblcity{
\begin{table}[!t]
\caption{\textbf{Cityscapes labels} $\leftrightarrow$ \textbf{Photograph.} We evaluate the generation quality for fake samples by mean square error (MSE) with ground-truth images.}
\centering\begin{tabular}{l|c|c}
\hline
Model & Labels $\rightarrow$ Photo & Photo $\rightarrow$ Labels \\\hline
CycleGAN & $0.728$ & $0.057$ \\
CR-CycleGAN & $0.625$ & $0.065$ \\
ACCR-CycleGAN (Ours) &\mbox{\boldmath $0.561$} & \mbox{\boldmath $0.040$} \\
\hline
\hline
\end{tabular}
\label{tbl:city}
\end{table}
}

\def\tblda{
\begin{table}[!t]
\caption{\textbf{Comparison of different types of data augmentation.}  We experiment with 7 types of image augmentation: (1) randomly cropping images by a few pixels, (2) randomly rotating images by a few degrees, (3) combination of random cropping and random rotation, (4)  applying cutout~\cite{cutout}, (5)  applying random erasing~\cite{erasing}, (6) randomly changing the brightness, contrast, and saturation of the images, and (7) a combination of random cropping, rotation, and color jitter. All the models in the experiment are based on CycleGAN.}
\scalebox{0.92}[1]{
\centering\begin{tabular}{l|l|c|c}
\hline
Data Augmentation & Method & M $\rightarrow$ MM & MM $\rightarrow$ M \\\hline
\multirow{2}{*}{(1) Random Crop} & CR & $97.7 \pm 0.6$ &  $94.3 \pm 0.5$ \\
& ACCR (Ours) & \mbox{\boldmath $98.0 \pm 0.5$} & \mbox{\boldmath $94.5 \pm 0.5$} \\\hline
\multirow{2}{*}{(2) Random Rotation} & CR &  $97.9 \pm 0.3$ & $93.6 \pm 1.3$ \\
& ACCR (Ours)  & \mbox{\boldmath $98.1 \pm 0.1$} & \mbox{\boldmath $94.4 \pm 0.3$} \\\hline
\multirow{2}{*}{(3) Crop\&Rotation} & CR & $98.1 \pm 0.3$ & $92.7 \pm 1.1$ \\
& ACCR (Ours)  & \mbox{\boldmath $98.2 \pm 0.2$} & \mbox{\boldmath $94.3 \pm 0.2$} \\\hline
\multirow{2}{*}{(4) Cutout} & CR & $97.2 \pm 0.6$ & $93.1 \pm 1.3$ \\
& ACCR (Ours)  & \mbox{\boldmath $97.4 \pm 0.4$} & \mbox{\boldmath $94.0 \pm 0.9$} \\\hline
\multirow{2}{*}{(5) Random Erasing} & CR &  $96.6 \pm 1.2 $  & $92.5 \pm 1.5$ \\
& ACCR (Ours)  & \mbox{\boldmath $97.3 \pm 1.0$} & \mbox{\boldmath $93.6 \pm 0.8 $} \\\hline
\multirow{2}{*}{(6) Color Jitter} & CR &  $97.2 \pm 0.8$ & $95.0 \pm 0.3$ \\
& ACCR (Ours)  & \mbox{\boldmath $97.7 \pm 0.3$} & \mbox{\boldmath $95.2 \pm 0.1$} \\\hline
\multirow{2}{*}{(7) Crop\&Rotation\&Jitter} & CR &  $97.7 \pm 0.6$ &  \mbox{\boldmath $94.5 \pm 0.4$} \\
& ACCR (Ours)  & \mbox{\boldmath $97.8 \pm 0.4$} & \mbox{\boldmath $94.5 \pm 0.3$} \\
\hline 
\hline
\end{tabular}
\label{tbl:da}
}
\end{table}
}

\def\tblcyc{
\begin{table}[!t]
\caption{\textbf{Comparison of other cycle-consistent constraints.} ~\cite{acal} proposed Relaxed Cyclic Adversarial Learning (RCAL) as an extension of cycle-consistency~\cite{cyclegan}, which enforces feature-wise cycle-consistency by using task specific models.}
\centering\begin{tabular}{l|c|c}
\hline
Model & M $\rightarrow$ S & S $\rightarrow$ M \\\hline
RCAL & $68.2 \pm 10.9$ & $47.5 \pm 3.5$\\
CR-RCAL & $67.1 \pm 11.4$ & $55.7 \pm 1.4$\\
ACCR-RCAL (Ours) &  \mbox{\boldmath $72.0 \pm 8.0$} &  \mbox{\boldmath $57.8 \pm 9.4$} \\\hline
\hline
\end{tabular}
\label{tbl:cyc}
\end{table}
}

\def\tblspeed{
\begin{table}[!t]
\caption{\textbf{Training speed of discriminator updates.} We report the actual training speed of discriminator updates for CycleGAN on MNIST $\leftrightarrow$ MNIST-M with NVIDIA Tesla P100. GP denotes CycleGAN with Gradient Penalty~\cite{wgangp}.}
\centering\begin{tabular}{l|c|c|c|c}
\hline
Method & W/O & GP & CR & ACCR (Ours) \\\hline
Speed (step/s) & $1871\pm 14$ & $786 \pm 8$ & $1740 \pm 25$ & $1176 \pm 28$ \\
\hline 
\hline
\end{tabular}
\label{tbl:speed}
\end{table}
}

\def\figimg{
\begin{figure}
     \centering
     \begin{subfigure}[t]{0.235\textwidth}
         \centering
         \includegraphics[width=\textwidth]{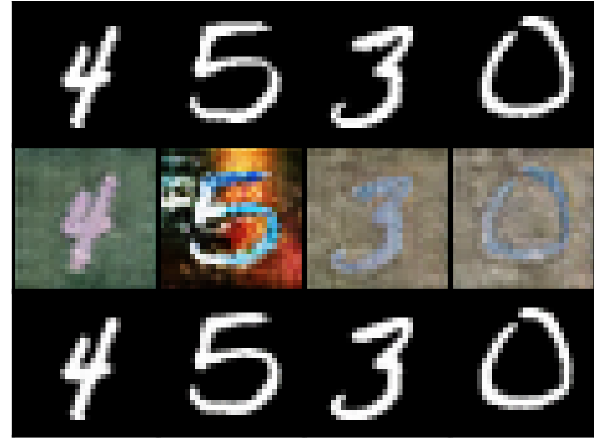}
         \caption{\textbf{MNIST $\rightarrow$ MNIST-M}}
         \label{fig:m_mm}
     \end{subfigure}
     \hfill
     \begin{subfigure}[t]{0.235\textwidth}
         \centering
         \includegraphics[width=\textwidth]{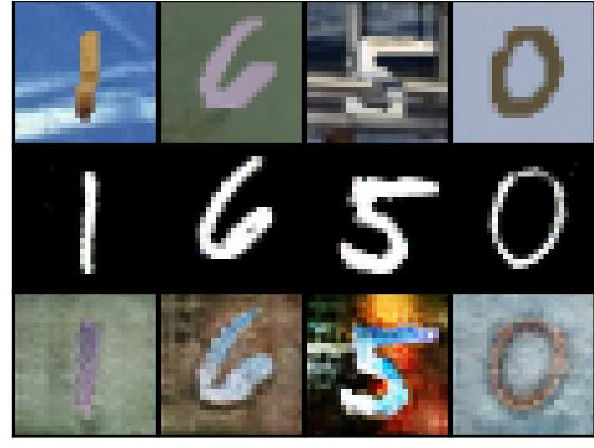}
         \caption{\textbf{MNIST-M $\rightarrow$ MNIST}}
         \label{fig:mm_m}
     \end{subfigure}
      \vfill
     \begin{subfigure}[t]{0.235\textwidth}
         \centering
         \includegraphics[width=\textwidth]{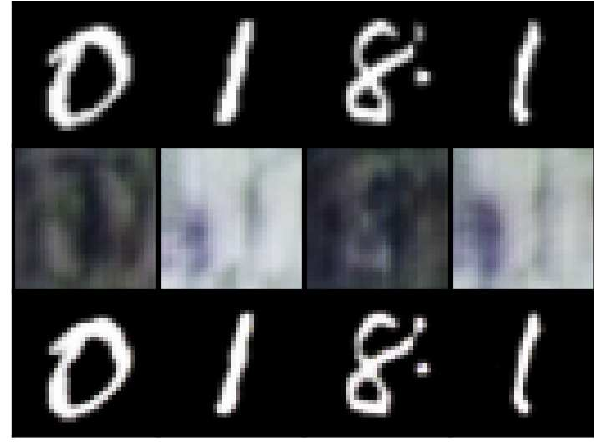}
         \caption{\textbf{MNIST $\rightarrow$ SVHN}}
         \label{fig:m_s}
     \end{subfigure}
     \hfill
     \begin{subfigure}[t]{0.235\textwidth}
         \centering
         \includegraphics[width=\textwidth]{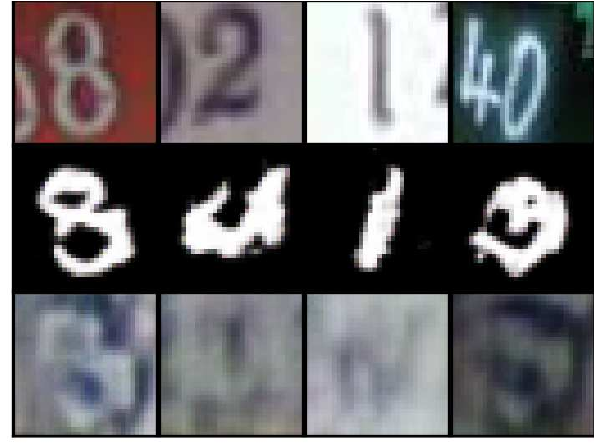}
         \caption{\textbf{SVHN $\rightarrow$ MNIST}}
         \label{fig:s_m}
     \end{subfigure}
        \caption{\textbf{Translation results of digit datasets.} In each translation result, Top: real samples, Middle: fake samples, Bottom: reconstructed samples. Reconstructed images at (b) MNIST-M $\rightarrow$ MNIST have a high variety of backgrounds due to the one-to-many projection of unpaired I2I translation. ACCR uses these samples for regularization. In the SVNH translations, the model often fails to generate reliable samples because of a large domain shift.}
        \label{fig:gen_img}
\end{figure}
}

%% file: abstract.tex
\begin{abstract}
Unpaired image-to-image (I2I) translation has received considerable attention in pattern recognition and computer vision because of recent advancements in generative adversarial networks (GANs). However, due to the lack of explicit supervision, unpaired I2I models often fail to generate realistic images, especially in challenging datasets with different backgrounds and poses. Hence, stabilization is indispensable for GANs and applications of I2I translation. Herein, we propose Augmented Cyclic Consistency Regularization (ACCR), a novel regularization method for unpaired I2I translation. Our main idea is to enforce consistency regularization originating from semi-supervised learning on the discriminators leveraging real, fake, reconstructed, and augmented samples. We regularize the discriminators to output similar predictions when fed pairs of original and perturbed images. We qualitatively clarify why consistency regularization on fake and reconstructed samples works well. Quantitatively, our method outperforms the consistency regularized GAN (CR-GAN) in real-world translations and demonstrates efficacy against several data augmentation variants and cycle-consistent constraints.

\end{abstract}

%% file: introduction.tex
\section{Introduction}
\fig{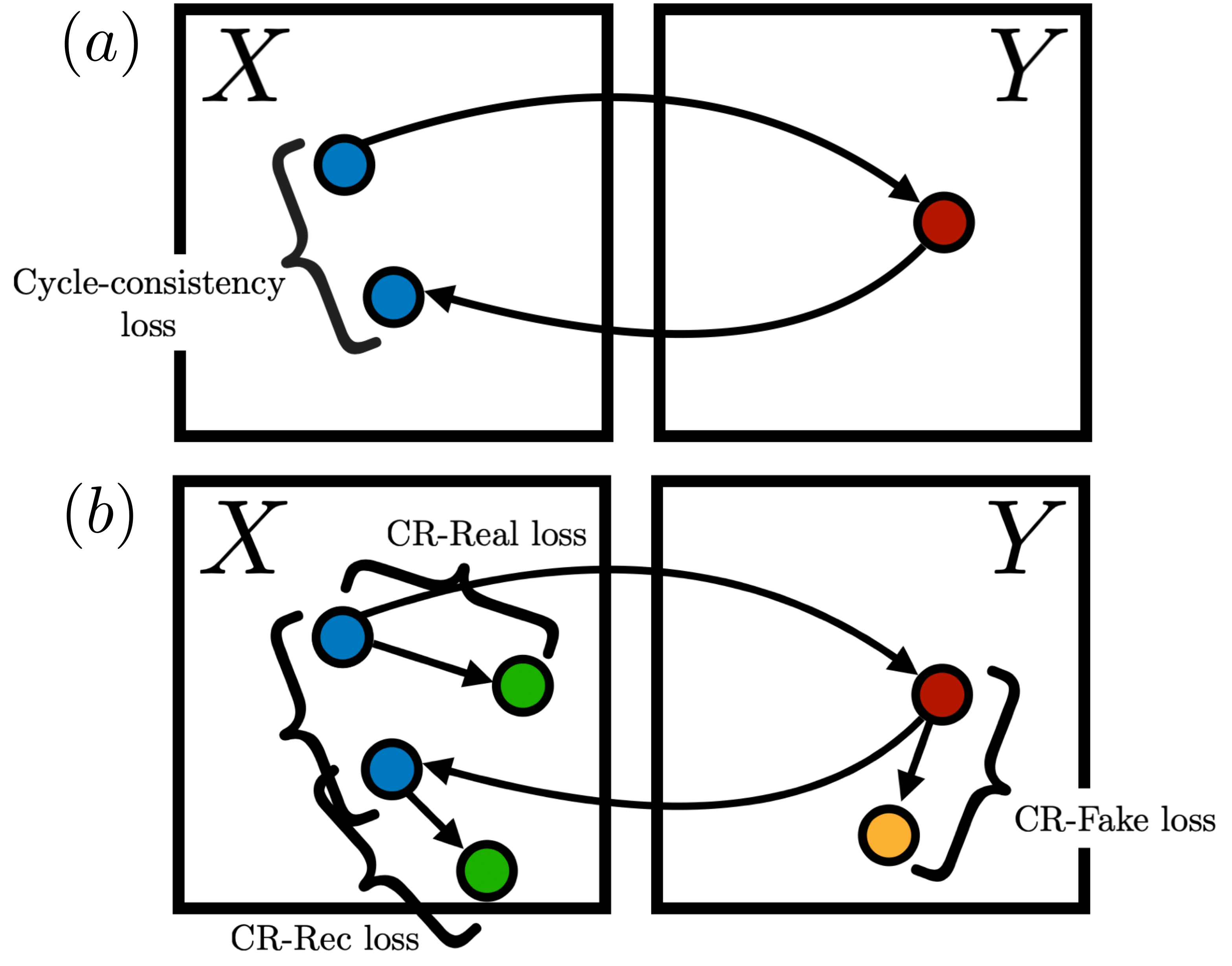}{\textbf{Illustration of our approach.}  (a) CycleGAN~\cite{cyclegan} contains two generators $G: X \mapsto Y$ and $F: Y \mapsto X$, and cycle-consistency loss, which enforces pixel-wise matching in the form of L1 loss between real and reconstructed data. 
 (b) Augmented Cyclic Consistency Regularization (ACCR) is an extension of consistency regularization on discriminators in unpaired I2I models leveraging each  real $x \in X$ and reconstructed $F(G(x)) \in X$ sample pair (\textbf{Blue}), and  augmented samples $T(x) \in X$, $T(F(G(x))) \in X$ (\textbf{Green}). $T(\cdot)$ denotes a semantics-preserving data augmentation function. A fake sample $G(x) \in Y$ is also employed (\textbf{Red}) translated from domain $X$ and the augmented sample  $T(G(x)) \in Y$ (\textbf{Yellow}). For simplicity, the other cycle is omitted.}
 
Image-to-image (I2I) translation aims to learn a function by mapping images from one domain to another. The I2I framework is applied to many tasks in the fields of machine learning and computer vision such as image-inpainting~\cite{contextencoder}, super-resolution~\cite{srcnn, srcnn2}, colorization~\cite{col}, style transfer~\cite{styletransfer, in, cbn, bin, adain}, domain adaptation~\cite{cycada,i2ida, sda, sbada, acal}, and person re-identification~\cite{camstyle, spgan, ptgan, ecn, dgnet}. We face challenges either in collecting aligned image pairs for training (e.g., \textit{summer} $\rightarrow$ \textit{winter}) or its inexistence (e.g., \textit{artwork} $\rightarrow$ \textit{photo}); thus, most work focuses on unpaired I2I models under the assumption that paired data are not available. However, trade-offs arise in training stability due to the absence of paired supervision. Even more problematic, the unpaired setting is based on ill-posed problems having infinitely many solutions and multimodal outputs where a single input may correspond to multiple possible outputs. To handle this, models employ complex and disentangled architectures~\cite{munit, egsc_it, drit}, which pose substantial difficulties from an optimization perspective.

In recent years, several variants of cycle-consistency constraints~\cite{acal}, normalization techniques~\cite{in,cbn,adain,bin,spade,ugatit}, and different latent space assumptions~\cite{unit,munit,drit,transgaga} have been investigated to achieve semantic-aware cycles, control style information, and disentangle features. Despite these advances, stabilization is rarely discussed because the issue is avoided by dealing with narrow translations or utilizing refined datasets with similar poses and backgrounds.  Furthermore, when it comes to real-world applications such as domain adaptation and person re-identification, data include various blurs, illuminations, or noise; therefore, training is not straightforward.

For GANs, several normalizations and gradient-based regularization techniques for the GAN discriminator have been studied such as batch normalization~\cite{bn}, layer normalization~\cite{ln}, spectral normalization~\cite{sngan}, and the gradient penalty~\cite{wgangp}. However, \cite{largestudygan} empirically revealed that simultaneous enforcement of both normalization and gradient-based regularization provides marginal gains or fails. Especially, I2I models adopt several normalization techniques~\cite{in,cbn,adain,bin,spade,ugatit} and accordingly I2I models with state-of-the-art gradient-based regularization are associated with more theoretical uncertainty than standard GANs.

Zhang et al.~\cite{crgan} first put forward consistency regularized GAN (CR-GAN) whereby consistency regularization was introduced to the GAN discriminator from semi-supervised learning. CR-GAN surpasses gradient-based approaches, but CR-GAN is limited to real samples and regularization failures can occur using generated images for standard GANs.

In this work, we propose augmented cyclic consistency regularization (ACCR), a novel regularization technique in unpaired I2I translation without gradient information which incorporates consistency regularization on the discriminators leveraging three types of samples: real, fake, and reconstructed images. We augment these data feeding to the discriminators and penalize sensitivity to perturbations. We show an intuitive illustration of our method in CycleGAN~\cite{cyclegan} in Fig.~\ref{eye_catch2}. 

Qualitatively, I2I models guarantee quality of both fake and reconstructed samples due to faster learning and lower potential of mode collapse. Thus, we justify the use of these images. Quantitatively, our method outperforms the CycleGAN baseline, the CR-GAN method, and models with consistency regularization using fake and reconstructed samples respectively on MNIST $\leftrightarrow$ MNIST-M, MNIST $\leftrightarrow$ SVHN, Maps $\leftrightarrow$ Aerial photo, and Labels $\leftrightarrow$ Photo. ACCR-CycleGAN improves the baseline by 0.3\% on MNIST $\rightarrow$ MNIST-M, 2.3\% on MNIST-M $\rightarrow$ MNIST, 3.9\% on MNIST $\rightarrow$ SVHN, and 3.7\% on SVHN $\rightarrow$ MNIST as measured by classification accuracy on fake samples. In real-world translations, ACCR-CycleGAN surpasses the baseline by 26.4\% on Maps $\rightarrow$ Photo, 22.9\% on Photo $\rightarrow$ Maps, 29.8\% on Photo $\rightarrow$ Labels as evaluated by mean squared error with ground-truth images. More importantly, even in real-world tasks, ACCR can always further improve the state-of-the-art CR technique. In extensive ablation studies, ACCR outperforms the CR-GAN method in other types of data augmentation and cycle-consistent constraints.

The contributions of our work are summarized as follows:
\begin{itemize}
    \item We propose a novel, simple, and effective training stabilizer in unpaired I2I translation using real, fake, and reconstructed samples.
    \item We qualitatively explain why consistency regularization employing fake and reconstructed samples performs well in unpaired I2I models.
    \item Our ACCR quantitatively outperforms the CycleGAN baseline and the CR-GAN method in several datasets, for various cycle-consistent constraints, and with several commonly used data augmentation techniques, as well as combinations thereof.
\end{itemize}

%% file: related_work.tex
\section{Related Work}

\subsection{Image-to-Image Translation}
To learn the mapping function with paired training data, Pix2Pix~\cite{pix2pix} applies conditional GANs using both a latent vector and the input image. The constraint is enforced by the ground truth labels or pairwise correspondence at the pixel-level. CycleGAN~\cite{cyclegan}, DiscoGAN~\cite{discogan}, and UNIT~\cite{unit} employ a cycle-consistency constraint to simultaneously learn a pair of forward and backward mappings between two domains given unpaired training data, which is conditioned solely on an input image and accordingly produce one single output.

To achieve multimodal generation, BicycleGAN~\cite{bicyclegan} injects noise into mappings between latent and target spaces to prevent mode collapse in the paired setting. In unpaired multimodal translations, augmented CycleGAN~\cite{augmentedcyclegan} also injects latent code in the generators, and concurrent work~\cite{munit, egsc_it, drit} adopts disentangled representations to produce diversified outputs. 
 
In the field of domain adaptation, the CycleGAN framework is applied in cycle-consistent adversarial domain adaptation models~\cite{cycada, sbada, acal} and I2I translation based domain adaptation~\cite{i2ida, sda} is designed for semantic segmentation of the target domain images. Currently, GAN-based domain adaptation is introduced in person re-identification for addressing challenges in real-world scenarios. CycleGAN-based approaches~\cite{camstyle, spgan, ptgan, ecn} are widely adopted to transfer pedestrian image styles from one domain to another. State-of-the-art DG-Net~\cite{dgnet} makes use of disentangled architecture to encode pedestrians in appearance and structure spaces for implausible person image generation.
 
However, despite the wide range of use cases, unpaired I2I translation is more difficult from an optimization perspective because of the lack of supervision in the form of paired examples. Moreover, the latest multimodal methods incorporate domain-specific and domain-invariant encoders~\cite{munit, egsc_it, drit, transgaga}. These approaches often fail when the amount of training data is limited, or domain characteristics differ significantly~\cite{drit}. It is problematic to learn separate latent spaces, larger networks, and unconditional generation where the latent vector can be simply mapped to a full-size image in contrast to the previous conditional cases. Therefore, our work mainly focuses on the stabilization of unpaired I2I translation.

In general, all the models share a problem whereby the generators cannot faithfully reconstruct the input images since I2I models are inherently one-to-many mappings. For instance, in the translation of \textit{semantic labels} $\rightarrow$ \textit{photo}, original colors, textures, and lighting are impossible to fully recover and stochastically vary because the details are lost in the label domain. This is also the case for all other translations such as \textit{map} $\leftrightarrow$ \textit{photo}, and \textit{summer} $\leftrightarrow$ \textit{winter}, as well as digits. In our work, we make use of this drawback for improving the diversity of fake and reconstructed images in consistency regularization.

\subsection{Consistency Regularization}
Consistency regularization was first proposed in the semi-supervised learning literature~\cite{semiregu,tempensemble}. The fundamental idea is simple: a classifier should output similar predictions for unlabeled examples even after they have been randomly perturbed. The random perturbations contain data augmentation~\cite{cutout,erasing}, stochastic regularization (e.g. Dropout~\cite{dropout}) ~\cite{semiregu,tempensemble}, and adversarial perturbations~\cite{vat2}. Analytically, consistency regularization enhances the smoothness of function prediction~\cite{imsat,vat2}.

CR-GAN~\cite{crgan} introduces consistency regularization in the GAN discriminator and improves state-of-the-art FID scores for conditional generation. In addition, CR-GAN outperforms gradient-based regularizers: Gradient Penalty~\cite{wgangp}, DRAGAN~\cite{dragan} and JS-Regularizer~\cite{jsregularizer}. However, the CR-GAN method on generated images often fails. We seek to explore this limitation and demonstrate the effectiveness of adding consistency regularization which employs both fake and reconstructed images.

%% file: proposal.tex
\section{Proposed Method}
\figw{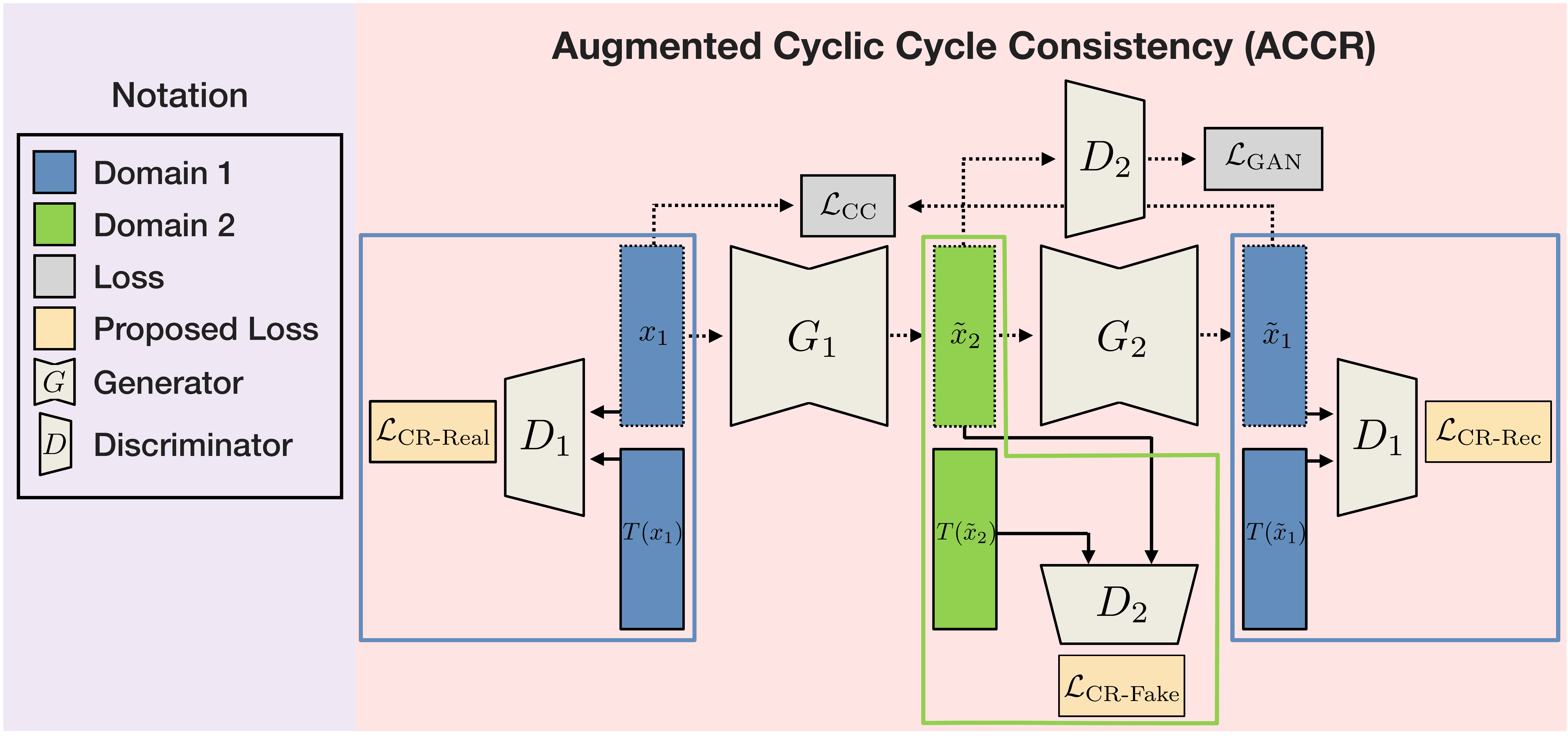}{\textbf{Overview of ACCR-CycleGAN.} CycleGAN~\cite{cyclegan} is depicted with Augmented Cyclic Consistency Regularization (ACCR), which consists of the CycleGAN architecture and consistency regularization losses on real, fake, and reconstructed images. The rectangles correspond to our proposed parts, $\mathcal{{L}}_{ \mathrm{CR \hyphen Real} }$, $\mathcal{{L}}_{\mathrm{CR \hyphen Fake}}$, and $\mathcal{{L}}_{\mathrm{CR \hyphen Rec}}$. $\mathcal{L}_\mathrm{GAN}$ and $\mathcal{L}_\mathrm{CC}$ indicate adversarial loss~\cite{gan} and cycle-consistency loss~\cite{cyclegan}, respectively. For expository purposes, the cycle in domain $X_2$ is omitted.}

\subsection{Preliminaries}
The goal of unpaired I2I translation is to learn the mapping within two domains $X_1$ and $X_2$ given training data $\{{x_1}_i\}_{i=1}^N$ and $\{{x_2}_i\}_{i=1}^M$ where ${x_1}_i \in X_1$ and ${x_2}_i \in X_2$.
We denote data distributions in two domains $x_1 \sim p_{data} (x_1)$, $x_2 \sim p_{data} (x_2)$, generators $G_1: X_1 \rightarrow X_2$, $G_2: X_2 \rightarrow X_1$, and discriminators $D_1$, $D_2$, where $D_1$ learns to distinguish real data $\{x_1\}$ from fake data $\{G_2 (x_2)\}$, $D_2$ learns differences $\{x_2\}$ from $\{G_1 (x_1)\}$. The objective consists of an adversarial loss~\cite{gan} and a constraint term $\mathcal{C}$ to encourage generators to produce samples that are structurally similar to inputs, and avoid excessive hallucinations and mode collapse that would increase the loss.

\subsubsection{Adversarial loss} The adversarial loss $\mathcal{L}_{\mathrm{GAN}}$ is employed to match the distribution of the fake images to the target image distribution, as written by
\setlength{\arraycolsep}{0.0em}
\begin{eqnarray}
    \mathcal{L}_{\mathrm{GAN}}\left (G_1, D_{2}\right) &{}={}& \mathbb{E}_{x_2 \sim p_{\mathrm{data}} (x_2)}\left[\log D_{2} (x_2)\right] \nonumber \\
    &{}+{}& \mathbb{E}_{x_1 \sim p_{\mathrm{data}} (x_1)}\left[\log \left (1-D_{2} (G_1  (x_1))\right)\right]. \nonumber \\
    \label{eq:gan}
\end{eqnarray}
\subsubsection{Unpaired I2I objective} Unpaired I2I translation requires the additional loss $\mathcal{C}$ to support forward and backward mappings between two domains. Thus, the full objective of unpaired I2I models (UI2I) is given by
\begin{eqnarray}
   \mathcal{L}_{\mathrm{UI2I}}\left (G_1, G_2, D_{1}, D_{2}\right) &{}={}& \mathcal{L}_{\mathrm{GAN}}\left (G_1, D_{2}\right) + \mathcal{L}_{\mathrm{GAN}}\left (G_2, D_{1}\right) \nonumber \\
   &{}+{}& \mathcal{C} (G_1, G_2).
  \label{eq:i2i}
\end{eqnarray}
CycleGAN~\cite{cyclegan} imposes a pixel-wise constraint in the form of cycle-consistency loss~$\mathcal{L}_{\mathrm{CC}}$~\cite{cyclegan} as the  constraint term $\mathcal{C}$, 
\begin{eqnarray}
\mathcal{L}_{\mathrm{CC}} (G_1, G_2) &{}={}& \lambda_1 \mathbb{E}_{x_1 \sim p_{\mathrm{data}} (x_1)}\left[\|G_2 (G_1 (x_1))-x_1\|_{1}\right] \nonumber \\
&{}+{}& \lambda_2 \mathbb{E}_{x_2 \sim p_{\mathrm{data}} x_2)}\left[\|G_1 (G_2 (x_2))-x_2\|_{1}\right]. \nonumber \\
\end{eqnarray}

\subsection{Consistency Regularization for GANs}
CR-GAN~\cite{crgan} proposes a simple, effective and fast training stabilizer introducing consistency regularization on the GANs discriminator. Assuming that the decision of the discriminator should be invariant to any valid domain-specific data augmentations, the sensitivity of the discriminator is penalized to randomly augmented data. It improves FID scores in conditional generations. The consistency regularization loss for discriminator $D$ is given by
\begin{eqnarray}
\mathcal{L}_{\mathrm{CR \hyphen Real}} (D)=\mathbb{E}_{x \sim p_{\mathrm{data} (x)}}\left[ \|D (x)-D (T (x))\|^{2} \right],
\label{eq:crgan}
\end{eqnarray}
where $T (\cdot)$ denotes a stochastic data augmentation function, e.g., flipping the image horizontally or translating the image by a few pixels. However,~\cite{crgan} reports that an additional regularization using generated images is not always superior to the original CR-GAN method.

\subsection{Augmented Cyclic Consistency Regularization}
We propose augmented cyclic consistency regularization (ACCR) for stabilizing training in unpaired I2I models. ACCR enforces consistency regularization on discriminators leveraging real, fake, reconstructed, and augmented samples. The goal is to verify the effectiveness of consistency regularization, even where fake and reconstructed data are employed from datasets which include noise (e.g., SVHN or MNIST-M). An overview of ACCR-CycleGAN is shown in Fig.~\ref{model001}.

We define consistency regularization losses on discriminators $D_1$ and $D_2$ leveraging real, fake, and reconstructed data denoted by $\mathcal{{L}}_{\mathrm{CR \hyphen Real}}$, $\mathcal{{L}}_{\mathrm{CR \hyphen Fake}}$, and $\mathcal{{L}}_{\mathrm{CR \hyphen Rec}}$, respectively. $\mathcal{{L}}_{\mathrm{CR \hyphen Real}}$ which is identical to CR-GAN is written as
\begin{eqnarray}
\lefteqn{\mathcal{{L}}_{\mathrm{CR \hyphen Real}} (D_1, D_2)} &{} {}& \nonumber \\
&{}={}&  \mathbb{E}_{x_1 \sim p_{\mathrm {data}} (x_1)} \left[ \|D_1 (x_1)-D_1 (T (x_1))\|^{2} \right] \nonumber \\
&{}+{}&  \mathbb{E}_{x_2 \sim p_{\mathrm {data}} (x_2)} \left[ \|D_2 (x_2)-D_2 (T (x_2))\|^{2} \right].
\label{eq:crreal}
\end{eqnarray}
Given fake samples \{$G_1 (x_1)$\}, \{$G_2 (x_2)$\} and augmented samples \{$T(G_1 (x_1))$\},  \{$T(G_2 (x_2))$\}, $\mathcal{{L}}_{\mathrm{CR-Fake}}$ is written as
\begin{eqnarray}
\lefteqn{\mathcal{{L}}_{\mathrm{CR \hyphen Fake}}(D_1, D_2)} \nonumber \\
 &{}={}& \mathbb{E}_{x_1 \sim p_{\mathrm {data}} (x_1)} \left[ \|D_2 (G_1 (x_1))-D_2 (T (G_1 (x_1)))\|^{2}\right] \nonumber \\
&{}+{}& \mathbb{E}_{x_2 \sim p_{\mathrm {data}} (x_2)} \left[ \|D_1 (G_2 (x_2))-D_1 (T (G_2 (x_2)))\|^{2} \right], \nonumber \\
\label{eq:crfake}
\end{eqnarray}
where $T (\cdot)$ denotes a stochastic data augmentation function which is semantics-preserving such as random crop, random rotation, or cutout~\cite{cutout}. Given reconstructed samples \{$G_2 (G_1 (x_1))$\},  \{$G_1 (G_2 (x_2))$\} and augmented samples \{$T(G_2 (G_1 (x_1)))$\},  \{$T(G_1 (G_2 (x_2)))$\}, $\mathcal{{L}}_{\mathrm{CR-Rec}}$ is written as
\begin{eqnarray}
\lefteqn{\mathcal{{L}}_{\mathrm{CR \hyphen Rec}} (D_1, D_2)} \nonumber \\
  &{}={}& \mathbb{E}_{x_1 \sim p_{\mathrm {data}} (x_1)} [ \|D_1 (G_2 (G_1 (x_1))) \nonumber \\
  &{}{}&\quad \quad \quad \quad \quad \quad \quad -\ D_1 (T (G_2 (G_1 (x_1))))\|^{2} ] \nonumber \\
  &{}+{}& \mathbb{E}_{x_2 \sim p_{\mathrm {data}} (x_2)} [ \|D_2 (G_1 (G_2 (x_2))) \nonumber \\
  &{}{}&\quad \quad \quad \quad \quad \quad \quad -\ D_2 (T (G_1 (G_2 (x_2))))\|^{2} ].
\label{eq:crrec}
\end{eqnarray}
By default, we use the random crop as $T (\cdot)$, and explore the effects of other functions in Section~\ref{sec:abl:da}. 

\subsection{Full Objective}
Finally, the objective of augmented cyclic consistency regularized unpaired I2I models (ACCR-UI2I) given unpaired data is written as
\begin{eqnarray}
\mathcal{L}_{\mathrm{UI2I}}^{\mathrm{ACCR}}\left (G_{1}, G_{2}, D_{1}, D_{2}\right)  &{}={}&\mathcal{L}_{\mathrm{GAN}}\left (G_{1}, D_{2}\right) \nonumber \\
&{}+{}& \mathcal{L}_{\mathrm{GAN}}\left (G_{2}, D_{1}\right) \nonumber \\
&{}+{}& \mathcal{C}\left (G_{1}, G_{2}\right) \nonumber \\
&{}+{}& \lambda_{\mathrm{Real}}   (\mathcal{{L}}_{\mathrm{CR \hyphen Real}} (D_1, D_2)) \nonumber \\
&{}+{}& \lambda_{\mathrm{Fake}}  (\mathcal{{L}}_{\mathrm{CR \hyphen Fake}} (D_1, D_2)) \nonumber \\
&{}+{}& \lambda_{\mathrm{Rec}}  (\mathcal{{L}}_{\mathrm{CR \hyphen Rec}} (D_1, D_2)),
\label{eq:accr3}
\end{eqnarray}
where the hyper-parameters $\lambda_{\mathrm{Real}}$, $\lambda_{\mathrm{Fake}}$, and $\lambda_{\mathrm{Rec}}$ control the weights of the regularization terms.

\tbldata
\tblmap

%% file: experiments.tex
\section{Experiments}
This section validates our proposed ACCR method in digit translations with various backgrounds (MNIST $\leftrightarrow$ MNIST-M and MNIST $\leftrightarrow$ SVHN) and real-world translations (Maps $\leftrightarrow$ Aerial photo and Cityscapes labels $\leftrightarrow$ Photo). First, we present details concerning the datasets and experimental implementation. Next, we conduct a quantitative analysis to demonstrate the performance on the translations and investigate the feature distance between real and augmented samples in discriminators for verifying the effect of ACCR. We then conduct a qualitative analysis for the generated samples and failure cases of the CR-GAN. Finally, we conduct ablation studies to compare consistency regularization utilizing fake and reconstructed images and explore the importance of choices with respect to data augmentation and cycle-consistent constraints.

\subsection{Datasets}\label{sec:data}
\textbf{MNIST (M) $\leftrightarrow$ MNIST-M (MM):}
MNIST~\cite{mnist} contains centered,
28$\times$28 pixel grayscale images of single-digit numbers
on a black background, 60,000 images for training and 10,000 for validation. We rescale to 32$\times$32 pixels and extend the channel to RGB. MNIST-M~\cite{mnist_m} contains centered, 32$\times$32 pixel digits on a variant background which is substituted by a randomly extracted patch obtained from color photos from BSDS500~\cite{bsds500}, 59,000 images for training and 1,000 for validation.

\textbf{MNIST (M) $\leftrightarrow$ SVHN (S):} We preprocess MNIST~\cite{mnist} as above. SVHN~\cite{svhn} is the challenging real-world Street View House Number dataset, much larger in scale than the other considered datasets. It contains 32$\times$32 pixel color samples, 73,257 images for training and 26,032 images for validation. Besides varying the shape and texture, the images often contain extraneous numbers in addition to those which are labeled and centered. 

\textbf{Google Aerial Photo $\leftrightarrow$ Maps:} The dataset consists of 3292 pairs of aerial photos and corresponding maps. In our experiments,  we resize the images from 600$\times$600 pixels to 256$\times$256 pixels. We use 1098 images for training and 1096 images for testing.

\textbf{Cityscapes Labels $\leftrightarrow$ Photo:} The dataset contains 2975 pairs of training images from the Cityscapes training set~\cite{city} and its segmentation labels.  In our experiments,  we resize the images to 256$\times$256 pixels. We use the Cityscapes validation set for testing.

\subsection{Implementation}\label{sec:imp}

\subsubsection{Network architecture}
We adopt architecture for our networks based on Hoffman et al~\cite{cycada}. The generator consists of two slide-2 convolutional layers followed by two residual blocks and then two deconvolution layers with slide $\frac{1}{2}$. The discriminator network consists of PatchGAN~\cite{pix2pix} with 5 convolutional layers. For all digit experiments, we use a variant of LeNet~\cite{mnist} architecture with 2 convolutional layers and 2 fully connected layers for 32$\times$32 pixel images.

\subsubsection{Training details}
In terms of $\mathcal{L}_{\mathrm{GAN}}$, we replace binary cross-entropy loss by a least-squares loss~\cite{lsgan} to stabilize GANs optimization as per~\cite{cyclegan}. For digit experiments, we exploit the Adam solver~\cite{adam} to optimize the objective with a learning rate of 0.0002 on the generators and 0.0001 on the discriminators, and first (second) moment estimates of  0.5 (0.999). We train for the first 10 epochs and then linearly decay the learning rate to zero over 20 epochs. Moving on, $\lambda_{\mathrm{real}}$ is set to 1 and $\lambda_{\mathrm{fake}}$ and $\lambda_{\mathrm{rec}}$ linearly increase from zero to half of $\lambda_{\mathrm{real}}$ because higher quality and diversified samples are guaranteed in the latter part of the training. We set the magnitude of cycle-consistency $\lambda_{\mathrm{cyc}}$ as 10 in MNIST and 0.1 in MNIST-M and SVHN. For real-world translations (Maps $\leftrightarrow$ Aerial photo and Labels $\leftrightarrow$ Photo), we exploit the hyper-parameter setting of CycleGAN~\cite{cyclegan}. By default, random crop is adopted as the stochastic data augmentation function at digit experiments. Color jitter is utilized for the photo domains in the real-world translations (Maps $\leftrightarrow$ Aerial photo and Labels $\leftrightarrow$ Photo), since it preserves the geometric structure of maps and segmentation labels, and tends to occur in the real world situations.

\subsubsection{Evaluation details}
For evaluation of all digit translations, we train revised LeNets~\cite{mnist} in MNIST, MNIST-M, and SVHN, which reach classification accuracies of 99.2\%, 97.5\%, and 91.0\%, respectively. We fix these classifiers for the tests, experiment 5 times with different random seeds, and report classification accuracies (\%) on fake samples. In Maps $\leftrightarrow$ Aerial photo and Labels $\leftrightarrow$ Photo translations, we measured the quality and level of detail for fake samples by mean square error (MSE) with ground-truth images.

\tblcity
\subsection{Quantitative Analysis}\label{sec:quan}
Our proposed method is compared against CR-GAN~\cite{crgan} in Table~\ref{tbl:data}. We conduct experiments on CycleGAN~\cite{cyclegan} as a baseline, CR-CycleGAN, a CycleGAN with consistency regularization using real samples, and our ACCR-CycleGAN on MNIST $\leftrightarrow$ MNIST-M and MNIST $\leftrightarrow$ SVHN. ACCR-CycleGAN outperforms CycleGAN and CR-CycleGAN in digit translations. To confirm the statistical significance of the results, we conduct paired sample T-tests by comparing our proposal with the baseline. CR + CR-Fake, CR + CR-Rec, and ACCR demonstrate statistically significant differences at the level of significance $\alpha=0.05$ for the translations of MNIST-M $\rightarrow$ MNIST and SVHN $\leftrightarrow$ MNIST. The performance of the MNIST $\rightarrow$ MNIST-M seems saturated. Therefore, we next validate ACCR for the more complex tasks.

We conduct experiments on two real-world I2I tasks: Maps $\leftrightarrow$ Aerial photo and Labels $\leftrightarrow$ Photo on Cityscapes. The results are shown in Table~\ref{tbl:map}, Table~\ref{tbl:city}, Fig.~\ref{map}, and Fig.~\ref{city}. Although the performance of the three models is saturated at Photo $\rightarrow$ Maps (Table~\ref{tbl:map}), ACCR has the lowest (best) scores in all the other translations. Hence, ACCR is generic regularization for real-world problems.

To identify the sensitivity of the discriminators to the augmented data, we calculate the mean squared error (MSE) in the feature space between the real and augmented data as shown in Table~\ref{tbl:dist}. ACCR and CR decrease the distance to the baseline, and ACCR exerts a greater impact than the baseline and CR, especially in MNIST-M. Therefore, the discriminators of ACCR are robust to inputs' perturbation and give further semantics-aware feedback to the generators in adversarial training.

\tbldist
\figimg
\subsection{Qualitative Analysis} \label{sec:qual}
Unpaired I2I problems are innately ill-posed and thus could have infinite solutions. Here we show generated samples in Fig.~\ref{fig:gen_img}. It seems impossible to determine only one mapping from a grayscale to a color background in the translation from real to fake on MNIST $\rightarrow$ MNIST-M (Fig.~\ref{fig:m_mm}) and the reconstruction on MNIST-M $\rightarrow$ MNIST (Fig.~\ref{fig:mm_m}). However, we leverage the stochastic property as diversified samples of consistency regularization. Indeed, the meaningful regularization corresponds to either CR-Fake or CR-Rec. Therefore, the combination of both CR-Fake and CR-Rec (ACCR) is reasonable.

Furthermore, CR-GAN~\cite{crgan} reported that consistency regularization on generated samples (CR-Fake) does not always lead to improvements. By investigating this limitation, we found that standard GANs fail to produce recognizable samples at the initial and end steps because the GANs are unable to fully capture the data distribution and may cause mode collapse, respectively. However, unpaired I2I translation induces these problems to a lesser extent due to image conditioning and the constraint term $\mathcal{C}$.  Hence, I2I models can preserve semantics even at the first and end epochs and this justifies using fake and reconstructed images for consistency regularization.

\fig{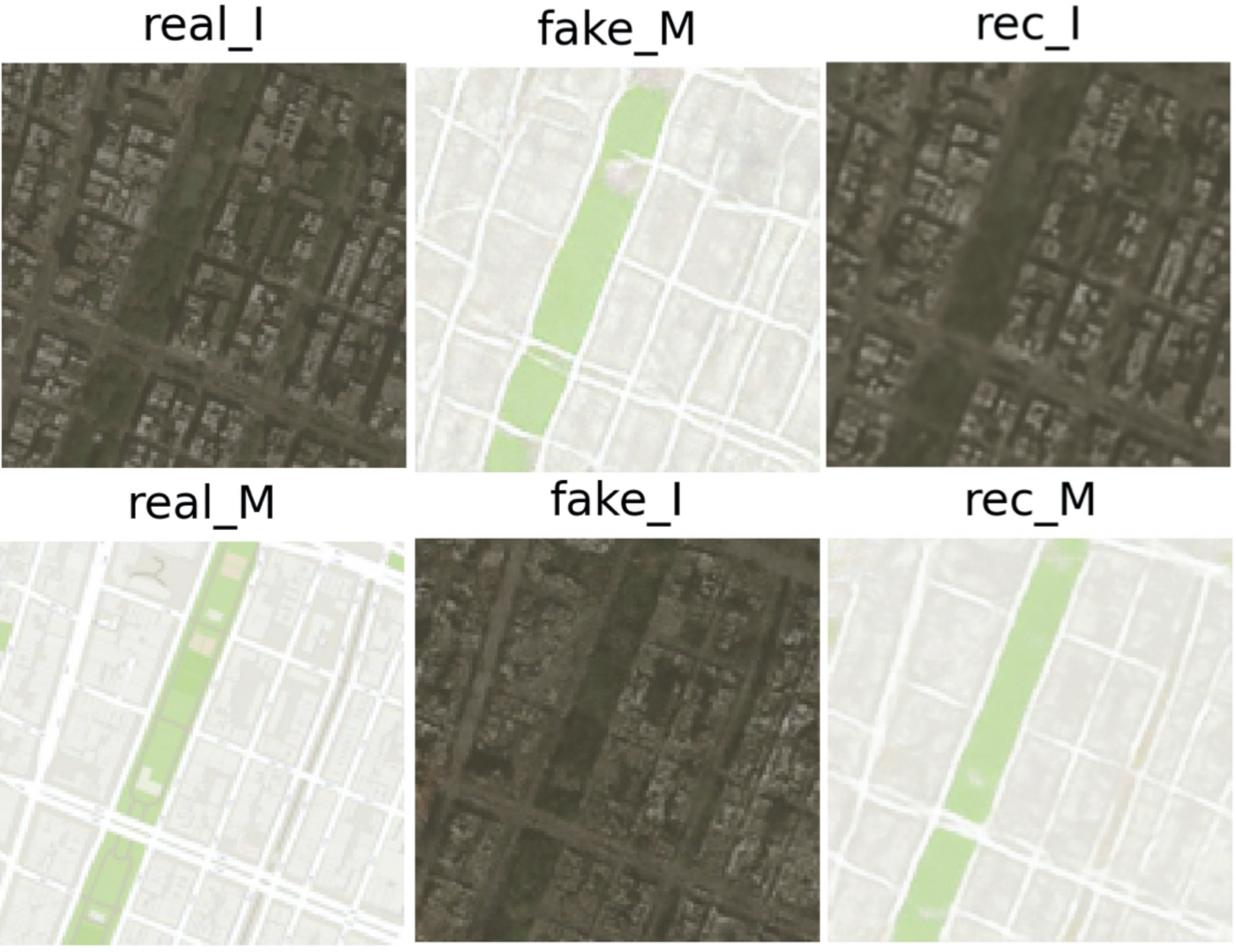}{\textbf{Translation results of Aerial Photo $\leftrightarrow$ Maps.} The dataset contains paired images. The ground-truth image of a fake map (\textit{fake\_M}) is \textit{real\_M}, and vice versa.}
\fig{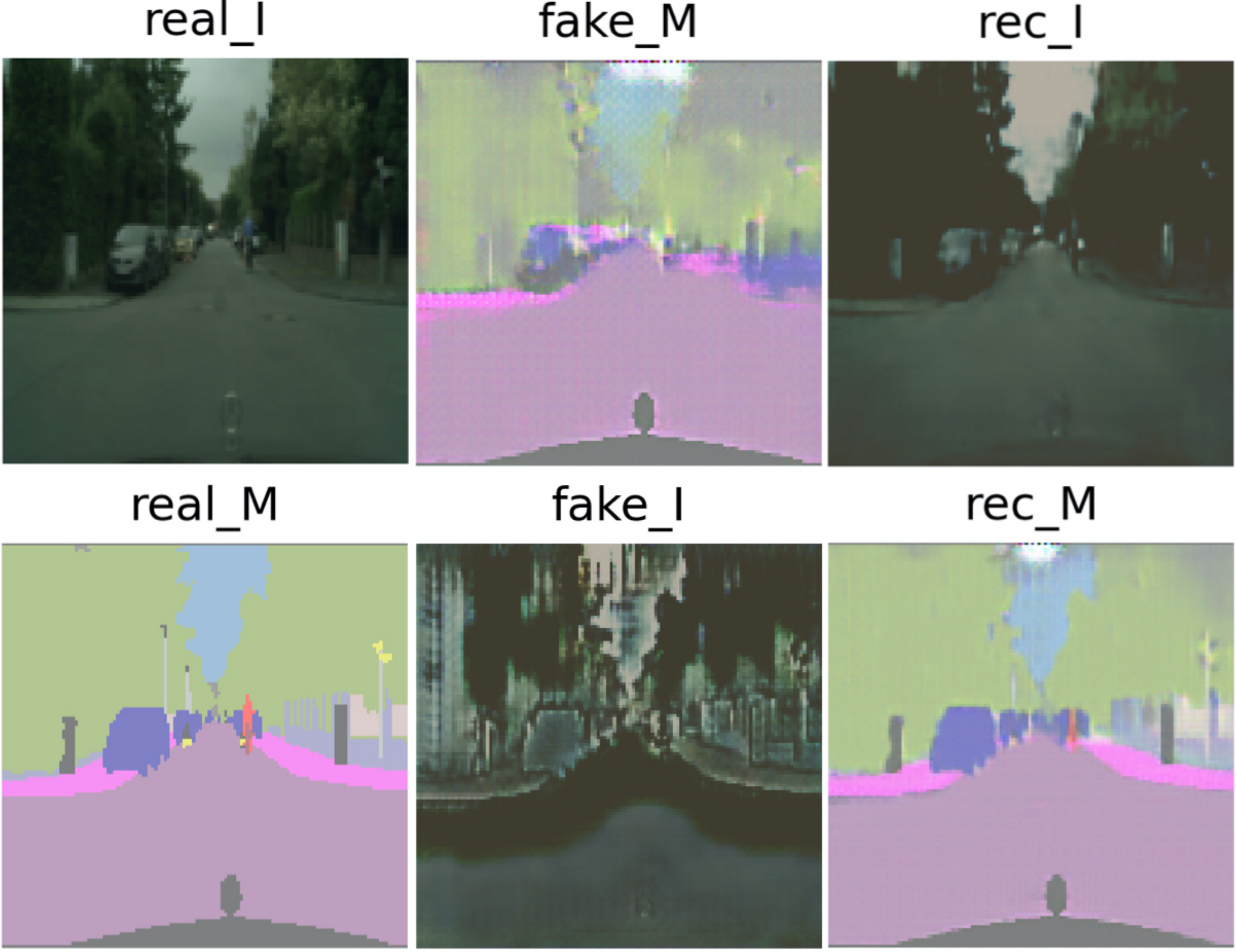}{\textbf{Translation results of Cityscapes Labels $\leftrightarrow$ Photo.} The dataset contains paired images. The ground-truth image of a fake mask (\textit{fake\_M}) is \textit{real\_M}, and vice versa.}

\subsection{Ablation Studies}
\subsubsection{Comparison with CR-Fake, CR-Rec, and ACCR}
To explore the effect of CR-Fake, CR-Rec, and ACCR, we compare each model on MNIST $\leftrightarrow$ MNIST-M and MNIST $\leftrightarrow$ SVHN as shown in Table~\ref{tbl:data}. As shown in Fig.~\ref{fig:gen_img}, the generation fails in a fake image of (c) MNIST $\rightarrow$ SVHN and a reconstructed image of (d) SVHN $\rightarrow$ MNIST due to complexity of the SVHN domain. Thus, CR + CR-Fake in MNIST $\rightarrow$ SVHN and CR + CR-Rec in SVHN $\rightarrow$ MNIST do not improve so much. Although CR-Fake and CR-Rec are sometimes inferior to CR, but ACCR always achieves the best. The gains of ACCR are discussed in Section~\ref{sec:qual}.
\tblda
\tblcyc
\subsubsection{Comparison with other data augmentation} \label{sec:abl:da}We compare several augmentation techniques in semantics-preserving ways (i.e., random crop, random rotation, cutout~\cite{cutout}, random erasing~\cite{erasing}, color jitter, and combinations thereof), as shown in Table~\ref{tbl:da}. ACCR tends to outperform the CR-GAN method in commonly used data augmentation techniques.
\subsubsection{Comparison with other cycle-consist constraints} \label{sec:abl:cyc}
Table~\ref{tbl:cyc} also shows results of an experiment with CycleGAN with Relaxed Cyclic Adversarial Learning (RCAL), which is a much looser constraint than having consistency in the pixel space, to verify our regularization with feature-level cycle-consistent constraints. RCAL is a naive extension of CycleGAN to the semantic-aware cycles using task-specific classifiers. ACCR-RCAL surpasses the RCAL baseline and CR-RCAL. Therefore, ACCR does not limit the choice of the constraint in pixel space. Rather, it is compatible with feature-wise cycle-consistent models.
\subsubsection{Training speed} \label{sec:abl:speed}In terms of computational cost, we measure the actual update speeds of the discriminators for ACCR-CycleGAN with NVIDIA Tesla P100 in Table~\ref{tbl:speed}.  ACCR marginally increases the forward pass of the discriminators compared with CR. ACCR-CycleGAN is around 1.5 times faster than CycleGAN with Gradient Penalty~\cite{wgangp}. We observe that CycleGAN with Gradient Penalty sometimes degrades from the baseline as observed in \cite{largestudygan, crgan}.
\subsubsection{Lambda sensitivity} \label{sec:abl:sensi}An experimental limitation of CR-CycleGAN and ACCR-CycleGAN lies in a lack of robustness for different lambda hyper-parameters. We conduct experiments for different hyper-parameters of CycleGAN and CR-CycleGAN where $\lambda_{\mathrm{Real}}$ ranges from 0.1 to 100. The results are shown in Fig~\ref{cyc_vs_cr_cyc2}. Due to the complexity of dual GANs optimization and original regularization terms of I2I models (e.g., consistency regularization), we cannot find the robustness in I2I translation as the significant results that CR-GAN reported~\cite{crgan}, especially when $\lambda_{\mathrm{Real}}$ is large. For ACCR-CycleGAN, we fixed the hyper-parameters that recorded the best in CR-CycleGAN. By the observation that real samples are more reliable than fake and reconstructed data for consistency regularization, we set  $\lambda_{\mathrm{Real}} > \lambda_{\mathrm{Fake}}$ = $\lambda_{\mathrm{Rec}}$ rather than $\lambda_{\mathrm{Real}} = \lambda_{\mathrm{Fake}}$ = $\lambda_{\mathrm{Rec}}$. This adjustment is also workable in real-world datasets.

\tblspeed
\fig{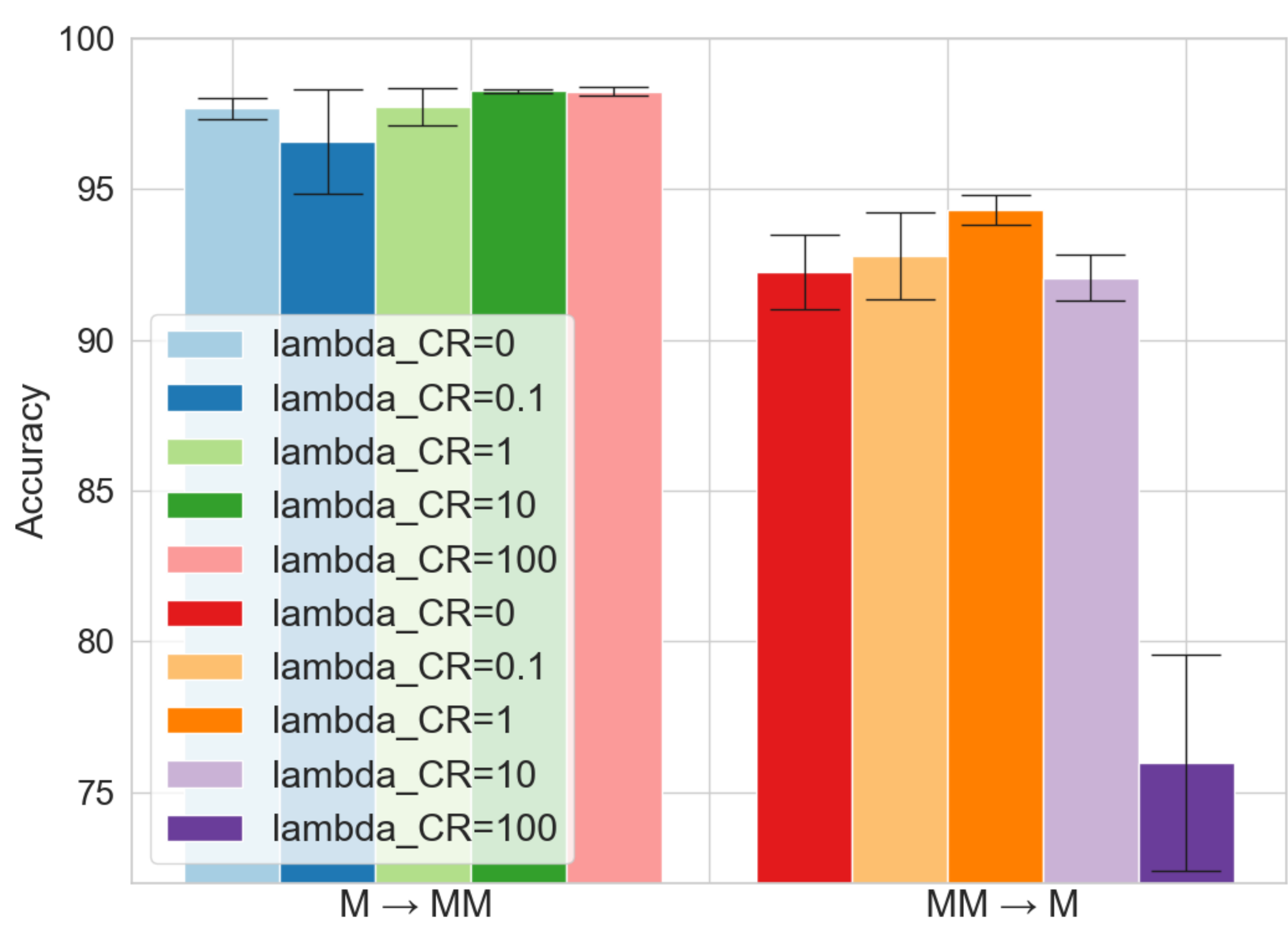}{\textbf{Different choices of $\lambda_{\mathrm{Real}}$.} lambda\_CR stands for the value of $\lambda_{\mathrm{Real}}$. The bars of $\lambda_{\mathrm{Real}}=0$ denote the CycleGAN baseline and the others are CR-CycleGANs.}

%% file: conclusion.tex
\section{Conclusion}
In this paper, we propose a novel, simple, and effective training stabilizer ACCR in unpaired I2I translation. We demonstrate the effectiveness of adding consistency regularization using both fake and reconstructed data. In experiments, our ACCR outperforms the baseline and the CR-GAN method in several digit translations and real-world translations. Furthermore, the proposed method surpasses the CR-GAN in various situations where the cycle-consistent constraint and the data augmentation function are different.

\newpage
{\noindent\textbf{Acknowledgments:}} We thank Takuma Yagi and Hiroaki Aizawa for comments on an earlier version of the manuscript. This work was partially supported by JSPS KAKENHI Grant Number 19K22865 and a hardware donation from Yu Darvish, a Japanese professional baseball player for the Chicago Cubs of Major League Baseball.